\documentclass[11pt,a4paper]{article}
\usepackage[hyperref]{emnlp2020}
\usepackage{times}
\usepackage{latexsym}

\usepackage{microtype}

\aclfinalcopy % Uncomment this line for the final submission
\usepackage[normalem]{ulem}
\usepackage{multirow}
\usepackage{makecell}
\usepackage{soul}
\usepackage{enumitem}
\usepackage{adjustbox}
\usepackage{amsmath}
\usepackage{amssymb}
\usepackage{array}
\usepackage{graphicx}
\usepackage{comment}
\usepackage{booktabs}
\usepackage{xcolor,colortbl} 
\usepackage{algorithm}
\usepackage{algorithmic}
\DeclareMathOperator*{\argmax}{arg\,max}

\usepackage{bbm}
\usepackage{tabularx}
\usepackage{hyperref}
\newcommand{\real}[1]{{\mathbb{R}^{{#1}}}}
\usepackage{soul}
\usepackage{tablefootnote}
\usepackage[normalem]{ulem}
\usepackage{multirow}
\usepackage{makecell}
\usepackage{soul}
\usepackage{enumitem}
\usepackage{adjustbox}
\usepackage{amsmath}
\usepackage{amssymb}
\usepackage{array}
\usepackage{graphicx}
\usepackage{comment}
\usepackage{booktabs}
\usepackage{xcolor,colortbl} 
\usepackage{algorithm}
\usepackage{algorithmic}
\usepackage{bbm}
\usepackage{tabularx}
\usepackage{hyperref}
\usepackage{graphicx}
\usepackage{subcaption}
\usepackage{arydshln}
\usepackage{array}
\usepackage{booktabs}

\title{
\textsc{Discern}: Discourse-Aware Entailment Reasoning Network for Conversational Machine Reading 
}

\author{\makecell{
Yifan Gao$^\dag$, Chien-Sheng Wu$^\ddag$, Jingjing Li$^\dag$, Shafiq Joty$^{\ddag,\S}$, \\ Steven C.H. Hoi$^\ddag$,
Caiming Xiong$^\ddag$, Irwin King$^\dag$, and Michael R. Lyu$^\dag$ }
\\
	{$^\dag$ The Chinese University of Hong Kong}\\
	{$^\ddag$ Salesforce Research} \quad $^\S$ Nanyang Technological University\\
    \tt{ $^\dag$\{yfgao,lijj,king,lyu\}@cse.cuhk.edu.hk}\\
	\tt{ $^\ddag$\{cswu,sjoty,shoi,cxiong\}@salesforce.com}
}

\date{}

\begin{document}
% Mark sections of captions for referring to divisions of figures
\newcommand{\tocite}[1]{{[\hl{CITE: #1]}}}
\newcommand{\todo}[1]{{[\hl{TODO: #1}]}}
\newcommand{\yifan}[1]{{[\hl{Yifan: #1}]}}

\newcommand{\modelname}{{Discourse-Aware Entailment Reasoning Network }}
\newcommand{\modelnamecap}{{\textbf{Disc}ourse-Aware \textbf{E}ntailment \textbf{R}easoning \textbf{N}etwork }}
\newcommand{\modelnameshort}{\textsc{Discern }}
\newcommand{\modelnameshortnsp}{\textsc{Discern}}
\newcommand{\sota}{{state-of-the-art }}
\maketitle
\begin{abstract}

Document interpretation and dialog understanding are the two major challenges for conversational machine reading.
In this work, we propose \modelnameshortnsp, a discourse-aware entailment reasoning network to strengthen the connection and enhance the understanding for both document and dialog. Specifically, we split the document into clause-like elementary discourse units (EDU) using a pre-trained discourse segmentation model, and we train our model in a weakly-supervised manner to predict whether each EDU is entailed by the user feedback in a conversation.
Based on the learned EDU and entailment representations, we either reply to the user our final decision ``yes/no/irrelevant" of the initial question, or generate a follow-up question to inquiry more information.
Our experiments on the ShARC benchmark (blind, held-out test set) show that \modelnameshort achieves \sota results of 78.3\% macro-averaged accuracy on decision making and 64.0 BLEU1 on follow-up question generation.
Code and models are released at \url{https://github.com/Yifan-Gao/Discern}.

\end{abstract}

\section{Introduction}

Conversational Machine Reading (CMR) is challenging because the rule text may not contain the literal answer, but provide a procedure to derive it through interactions \cite{saeidi-etal-2018-interpretation}.
In this case, the machine needs to read the rule text, interpret the user scenario, clarify the unknown user's background by asking questions, and derive the final answer.
Taking Figure \ref{fig.example} as an example, to answer the user whether he is suitable for the loan program, the machine needs to interpret the rule text to know what are the requirements, understand he meets ``American small business'' from the user scenario, ask follow-up clarification questions about ``for-profit business'' and ``not get financing from other resources'', and finally it concludes the answer ``Yes'' to the user's initial question.

\begin{figure}[t!]
\small
\begin{tabular}{p{0.95\columnwidth}}
\hline\hline
\textbf{Rule Text}: 7(a) loans are the most basic and most used type loan of the Small Business Administration's (SBA) business loan programs. It's name comes from section 7(a) of the Small Business Act, which authorizes the agency to provide business loans to \textbf{American small businesses}. The loan program is designed to assist \textbf{for-profit businesses} that are \textbf{not able to get other financing from other resources}. \\
\hline
\textbf{User Scenario}: I am a 34 year old man from the United States who owns their own business. We are an American small business. \\
\textbf{User Question}: Is the 7(a) loan program for me? \\
\textbf{Follow-up} $\mathbf{Q}_1$: Are you a for-profit business? \\
\textbf{Follow-up} $\mathbf{A}_1$: Yes. \\
\textbf{Follow-up} $\mathbf{Q}_2$: Are you able to get financing from other resources? \\
\textbf{Follow-up} $\mathbf{A}_2$: No. \\
\textbf{Final Answer}: Yes. (You can apply the loan.)\\
\hline\hline                
\end{tabular}
\caption{An example dialog from the ShARC \cite{saeidi-etal-2018-interpretation} dataset. The machine answers the user question by reading the rule text, interpreting the user scenario, and keeping asking follow-up questions to clarify the user's background until it concludes a final answer. Requirements in the rule text are bold.}
\label{fig.example}
\end{figure}

Existing approaches \cite{zhong-zettlemoyer-2019-e3,Sharma2019NeuralCQ,gao-etal-2020-explicit} decompose this problem into two sub-tasks. 
Given the rule text, user question, user scenario, and dialog history (if any), the first sub-task is to make a decision among ``Yes'', ``No'', ``Inquire'' and ``Irrelevant''.
The ``Yes/No'' directly answers the user question and ``Irrelevant'' means the user question is unanswerable by the rule text.
If the user-provided information (user scenario, previous dialogs) are not enough to determine his fulfillment or eligibility, an ``Inquire'' decision is made and the second sub-task is activated.
The second sub-task is to capture the underspecified condition from the rule text and generate a follow-up question to clarify it.
\citet{zhong-zettlemoyer-2019-e3} adopt BERT \cite{devlin-etal-2019-bert} to reason out the decision, and propose an entailment-driven extracting and editing framework to extract a span from the rule text and edit it into the follow-up question. 
The current \sota model EMT \cite{gao-etal-2020-explicit} uses a Recurrent Entity Network \cite{DBLP:conf/iclr/HenaffWSBL17} with explicit memory to track the fulfillment of rules at each dialog turn for decision making and question generation.

In this problem, document interpretation requires identification of conditions and determination of logical structures because rules can appear in the format of bullet points, in-line conditions, conjunctions, disjunctions, etc.
Hence, correctly interpreting rules is the first step towards decision making.
Another challenge is dialog understanding. The model needs to evaluate the user's fulfillment over the conditions, and jointly consider the fulfillment states and the logical structure of rules for decision making. For example, disjunctions and conjunctions of conditions have completely different requirements over the user's fulfillment states.
However, existing methods have not considered condition-level understanding and reasoning.

In this work, we propose \modelnameshortnsp: \modelnamecap.
To better understand the logical structure of a rule text and to extract conditions from it, we first segment the rule text into clause-like elementary discourse units (EDUs) using a pre-trained discourse segmentation model~\cite{DBLP:conf/ijcai/LiSJ18}.
Each EDU is treated as a condition of the rule text, and our model estimates its entailment confidence scores over three states: \textsc{Entailment}, \textsc{Contradiction} or \textsc{Neutral} by reading the user scenario description and existing dialog.
Then we map the scores to an entailment vector for each condition, and reason out the decision based on the entailment vectors and the logical structure of rules.
Compared to previous methods that do little entailment reasoning \cite{zhong-zettlemoyer-2019-e3} or use it as multi-task learning \cite{gao-etal-2020-explicit}, \modelnameshort is the first method to explicitly build the dependency between entailment states and decisions at each dialog turn.

\modelnameshort achieves new \sota results on the blind, held out test set of ShARC~\cite{saeidi-etal-2018-interpretation}.
In particular, \modelnameshort outperforms the previous best model EMT \cite{gao-etal-2020-explicit} by 3.8\% in micro-averaged decision accuracy and 3.5\% in macro-averaged decision accuracy.
Specifically, \modelnameshort performs well on simple in-line conditions and conjunctions of rules while still needing improvements on understanding disjunctions.
Finally, we conduct comprehensive analyses to unveil the limitation of \modelnameshort and current challenges for the ShARC benchmark.
We find one of the biggest bottlenecks is the user scenario interpretation, in which various types of reasoning are required.
% Code and models will be released to facilitate research along this line.

\section{\modelnameshort Model}

\begin{figure*}[t!]
\centering
\includegraphics[width=1.0\textwidth]{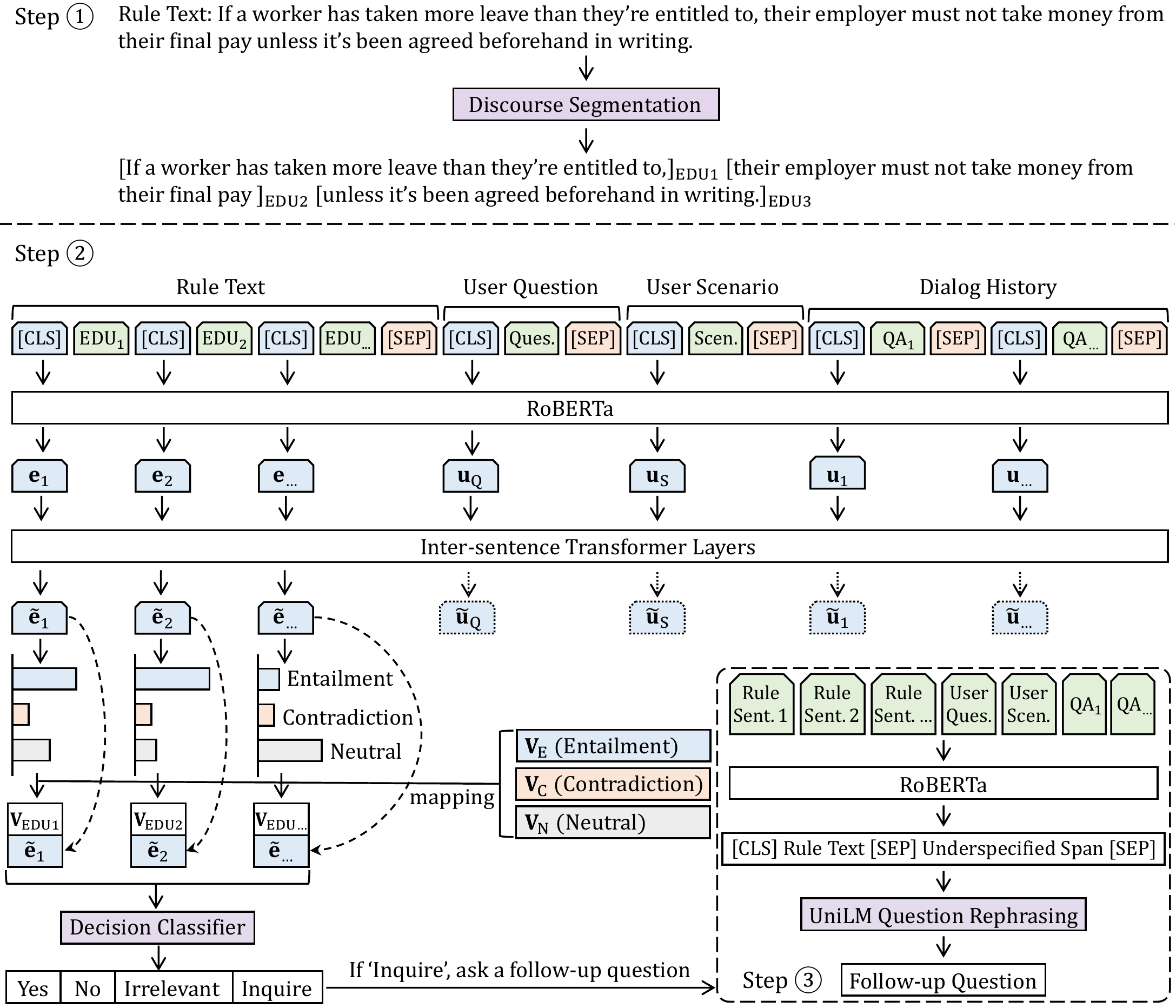}
\caption{The overall diagram of our proposed \modelnameshortnsp. \modelnameshort first segments the rule text into several elementary discourse units (EDUs) as conditions (Section \ref{sec.rule}). Then, taking the segmented conditions, user question, user scenario, and dialog history as inputs, \modelnameshort reasons out the decision among ``Yes'', ``No'', ``Irrelevant'' and ``Inquire'' (Section \ref{sec.dm}). If the decision is ``Inquire'', the question generation model asks a follow-up question (Section \ref{sec.qg}). \textit{(Best viewed in color)}
} 
\label{fig:model}
\end{figure*}

\modelnameshort answers the user question through a three-step process shown in Figure \ref{fig:model}:

\begin{enumerate}[leftmargin=*]
    \item First, \modelnameshort segments the rule text into individual conditions using discourse segmentation.
    \item Taking the user-provided information including the user question, user scenario and dialog history as inputs, \modelnameshort predicts the entailment state and maps it to an entailment vector for each segmented condition.
    Then it reasons out the decision by considering the logical structure of the rule text and the fulfillment of each condition.
    \item Finally, if the decision is ``Inquire'', \modelnameshort generates a follow-up question to clarify the underspecified condition in the rule text.
\end{enumerate}

\subsection{Rule Segmentation}\label{sec.rule}
The goal of rule segmentation is to understand the logical structure of the rule text and parse it into individual conditions for the ease of entailment reasoning. Ideally, each segmented unit should contain at most \textit{one} condition. Otherwise, it will be ambiguous to determine the entailment state for that unit.
Determining conditions is easy when they appear as bullet points, but in most cases (65\% samples in the ShARC dataset), one rule sentence may contain several in-line conditions as exemplified in Figure \ref{fig:model}.
To extract these in-line conditions, we find discourse segmentation in discourse parsing to be useful. 
In the Rhetorical Structure Theory or RST \cite{mann1988rhetorical} of discourse parsing, texts are first split into a sequence of clause-like units called elementary discourse units (EDUs).
We utilize an off-the-shelf discourse segmenter \cite{DBLP:conf/ijcai/LiSJ18}  to break the rule text into a sequence of EDUs. The segmenter uses a pointer network and achieves 92.2\% F-score with Glove vectors and 95.55\% F-score with ELMo embeddings  on the standard RST benchmark testset, which is close to human agreement of 98.3\% F-score \cite{joty-etal-2015-codra,lin-etal-2019-unified}. % is a pre-trained pointer model which achieves near human-level accuracy (92.2\% F-score) on the standard RST benchmark testset.
As exemplified in Figure \ref{fig:model} Step \textcircled{\raisebox{-0.9pt}{1}}, the rule sentence is broken into three EDUs, in which two conditions (``If a worker has taken more leave than they're entitled to'', ``unless it's been agreed beforehand in writing'') and the outcome (``their employer must not take money from their final pay'') are split out precisely.
% In average, there are 1.23 in-line conditions per rule sentence in the training data.
For rule texts which contain bullet points, we directly treat these bullet points as conditions.

\subsection{{Decision Making via Entailment Reasoning}}\label{sec.dm}
\paragraph{Encoding.}
As shown in Figure \ref{fig:model} Step \textcircled{\raisebox{-0.9pt}{2}}, 
inputs to \modelnameshort include the segmented conditions (EDUs) in the rule text, user question, user scenario, and follow-up question-answer pairs in dialog history, each of which is a sequence of tokens.
In order to get the sentence-level representations for all individual sequences, we insert an external \texttt{[CLS]} symbol at the start of each sequence, and add a \texttt{[SEP]} symbol at the end of every type of inputs.
Then, \modelnameshort concatenates all sequences together, and uses RoBERTa \cite{liu2019roberta} to encode the concatenated sequence. 
The encoded \texttt{[CLS]} token represents the sequence that follows it.
In this way, we extract sentence-level representations of conditions (EDUs) as $\mathbf e_1, \mathbf e_2, ..., \mathbf e_N$, and also the representations of the user question $\mathbf u_Q$, user scenario $\mathbf u_S$, and $M$ turns of dialog history $\mathbf u_1, ..., \mathbf u_M$.
All these vectorized representations are of $d$ dimensions (768 for RoBERTa-base).

\paragraph{Entailment Prediction.}
In order to reason out the correct decision for the user question, it is necessary to figure out the fulfillment of conditions in the rule text.
We propose to formulate the fulfillment prediction of conditions into a multi-sentence entailment task. 
Given a sequence of conditions (premises) and a sequence of user-provided information (hypotheses), a system should output \textsc{Entailment}, \textsc{Contradiction} or \textsc{Neutral} for each condition listed in the rule text. In this context, \textsc{Neutral} indicates that the condition has not been mentioned from the user information.

We utilize an inter-sentence transformer encoder \cite{transformer} to predict the entailment states for all conditions simultaneously.
Taking all sentence-level representations [$\mathbf e_1$; $\mathbf e_2$; ...; $\mathbf e_N$; $\mathbf u_Q$; $\mathbf u_S$; $\mathbf u_1$; ...; $\mathbf u_M$] as inputs, the $L$-layer transformer encoder makes each condition attend to all the user-provided information to predict whether the condition is entailed or not.
We also allow all conditions can attend to each other to understand the logical structure of the rule text.

Let the transformer encoder output of the $i$-th condition as $\tilde{\mathbf{e}}_i$,
we use a linear transformation to predict its entailment state:
\begin{align}
    \mathbf c_i &= \mathbf W_c \tilde{\mathbf{e}}_i + \mathbf b_c \in \real{3},
\end{align}
where $\mathbf c_i = [c_{\text{E},i}, c_{\text{C},i}, c_{\text{N},i}] \in \real{3}$ contains confidence scores of three entailment states \textsc{Entailment}, \textsc{Contradiction}, \textsc{Neutral} for the $i$-th condition in the rule text.

Since there are no ground truth entailment labels for individual conditions, we adopt a heuristic approach similar to \citet{gao-etal-2020-explicit} to get the noisy supervision signals.
Given the rule text, we first collect all associated follow-up questions in the dataset.
Each follow-up question is matched to a segmented condition (EDU) in the rule text which has the minimum edit distance.
For conditions in the rule text which are mentioned by follow-up questions in the dialogue history, we label the entailment state of a condition as \texttt{Entailment} if the answer for its mentioned follow-up question is \texttt{Yes}, and label the state of this condition as \texttt{Contradiction} if the answer is \texttt{No}.
% Then, for each sample in the training data, if a condition in the rule text has been mentioned by any follow-up question in the \textit{current} dialog, we take the \texttt{Yes}/\texttt{No} follow-up answer of that question as the entailment state \texttt{Entailment}/\texttt{Contradiction} of this condition.
The remaining conditions not covered by any follow-up question are labeled as \texttt{Neutral}.
Let $r$ indicate the correct entailment state. 
The entailment prediction is weakly supervised by the following cross entropy loss, normalized by total number of $K$ conditions in a batch:
\begin{align}\label{eqn:l_imp}
    \mathcal{L}_{\text{entail}} = - \frac{1}{K}\sum_{i=1}^K ~ \log~\text{softmax}(\mathbf c_i)_r
\end{align}

\paragraph{Decision Making.}
After knowing the entailment state for each condition in the rule text, the remaining challenge for decision making is to perform logical reasoning over different rule types such as disjunction, conjunction, and conjunction of disjunctions.
To achieve this, we first design three $d$-dimension entailment vectors $\mathbf{V}_{\text{E}}$ (Entailment), $\mathbf{V}_{\text{C}}$ (Contradiction), $\mathbf{V}_{\text{N}}$ (Neutral), and map the predicted entailment confidence scores of each condition to its vectorized entailment representation:
\begin{align}\label{eqn.cat}
    \mathbf{V}_{\text{EDU},i} &= \sum_{k \in [\text{E}, \text{C}, \text{N}]} c_{k,i} \mathbf{V}_{\text{k}} \in \real{d},
\end{align}
These entailment vectors are randomly initialized and then learned during training.
Finally, \modelnameshort jointly considers the logical structure of rules $\tilde{\mathbf{e}}_i$ and the entailment representations $\mathbf{V}_{\text{EDU},i}$ of conditions to make a decision:
\begin{align}
    \alpha_i &= \mathbf w_{\alpha}^\top [\mathbf{V}_{\text{EDU},i}; \tilde{\mathbf{e}}_i] + b_\alpha \in \real{1} \label{eqn:alpha} \\
    \tilde{\alpha}_i &= \text{softmax}(\mathbf{\alpha})_i \in [0,1]  \\
    \mathbf g &= \sum_i \tilde{\alpha_i} [\mathbf{V}_{\text{EDU},i}; \tilde{\mathbf{e}}_i] \in \real{2d} \label{eqn:summary} \\
    \mathbf z &= \mathbf W_z \mathbf g + \mathbf b_z \in \real{4}
\end{align}
where $[\mathbf{V}_{\text{EDU},i}; \tilde{\mathbf{e}}_i]$ denotes the vector concatenation, $\alpha_i$ is the attention weight for the $i$-th condition that determines whether the $i$-th condition should be taken into consideration for the final decision. 
$\mathbf z \in \real{4}$ contains the predicted scores for all four possible decisions ``Yes'', ``No'', ``Inquire'' and ``Irrelevant''.
Let $l$ indicate the correct decision, $\mathbf z$ is supervised by the following cross entropy loss:
\begin{align}\label{eqn:l_dec}
    \mathcal{L}_{\text{dec}} = -\log~\text{softmax}(\mathbf z)_l
\end{align}

\noindent The overall loss for the Step \textcircled{\raisebox{-0.9pt}{2}} decision making is the weighted-sum of decision loss and entailment prediction loss:
\begin{align}
    \mathcal{L} = \mathcal{L}_{\text{dec}} + \lambda \mathcal{L}_{\text{entail}} \label{eq:totalloss}
\end{align}

\subsection{Follow-up Question Generation}\label{sec.qg}
If the predicted decision is ``Inquire'', the follow-up question generation model is activated, as shown in Step \textcircled{\raisebox{-0.9pt}{3}} of Figure \ref{fig:model}. 
It extracts an \textit{underspecified} span from the rule text which is uncovered from the user's feedback, and rephrases it into a well-formed question.
Existing approaches put huge efforts in extracting the underspecified span, such as entailment-driven extracting and ranking \cite{zhong-zettlemoyer-2019-e3} or coarse-to-fine reasoning \cite{gao-etal-2020-explicit}.
However, we find that such sophisticated modelings may not be necessary, and we propose a simple but effective approach here.

We split the rule text into sentences and concatenate the rule sentences and user-provided information into a sequence.
Then we use RoBERTa to encode them into vectors grounded to tokens, as here we want to predict the position of a span within the rule text.
Let [$\mathbf{t}_{1,1}$, ..., $\mathbf{t}_{1,s_1}$; $\mathbf{t}_{2,1}$, ..., $\mathbf{t}_{2,s_2}$; ...; $\mathbf{t}_{N,1}$, ..., $\mathbf{t}_{N,s_N}$] be the encoded vectors for tokens from $N$ rule sentences, we follow the BERTQA approach \cite{devlin-etal-2019-bert} to learn a start vector $\mathbf w_s \in \real{d}$ and an end vector $\mathbf w_e \in \real{d}$ to locate the start and end positions, under the restriction that the start and end positions must belong to the same rule sentence:
\begin{align}\label{eqn:span}
    \text{Span} = \argmax_{i,j,k} (\mathbf{w}_s^\top \mathbf{t}_{k,i} + \mathbf{w}_e^\top \mathbf{t}_{k,j})
\end{align}
where $i,j$ denote the start and end positions of the selected span, and $k$ is the sentence which the span belongs to. The training objective is the sum of the log-likelihoods of the correct start and end positions.
To supervise the span extraction process, the noisy supervision of spans are generated by selecting the span which has the minimum edit distance with the to-be-asked question.
Lastly, following \citet{gao-etal-2020-explicit}, we concatenate the rule text and span as the input sequence, and finetune UniLM \cite{unilm}, a pre-trained language model to rephrase it into a question.

\begin{table*}
    \centering
    \small 
    \resizebox{0.95\textwidth}{!}{
    \begin{tabular}{@{~~~}l@{~} | @{~}c@{~} @{~}c@{~}  @{~~~~~}c@{~~~~} @{~~~~}c@{~~~~~} } 
    \Xhline{2\arrayrulewidth}
    \multirow{2}{*}{Models} &  \multicolumn{4}{c}{End-to-End Task (Leaderboard Performance)}  \\  
     & Micro Acc. & Macro Acc. & BLEU1 & BLEU4  \\
    \hline\hline
    Seq2Seq \cite{saeidi-etal-2018-interpretation} & 44.8 & 42.8 & 34.0 & {~}{~}7.8   \\
    Pipeline \cite{saeidi-etal-2018-interpretation}  & 61.9 & 68.9 & 54.4 & 34.4   \\
    BERTQA \cite{zhong-zettlemoyer-2019-e3}  & 63.6 & 70.8 & 46.2 & 36.3  \\
    UrcaNet \cite{Sharma2019NeuralCQ}   & 65.1 & 71.2 & 60.5 & 46.1   \\
    BiSon \cite{lawrence-etal-2019-attending}  & 66.9 & 71.6 & 58.8 & 44.3  \\
    E${}^{3}$ \cite{zhong-zettlemoyer-2019-e3}  & 67.6 & 73.3 & 54.1 & 38.7   \\
    EMT \cite{gao-etal-2020-explicit}  & {69.4} & {74.8} & {60.9} & {46.0}  \\
    EMT+entailment \cite{gao-etal-2020-explicit}  & {69.1} & {74.6} & {63.9} & \textbf{49.5}  \\
    \modelnameshort (our single model)  & \textbf{73.2} & \textbf{78.3} & \textbf{64.0} & {49.1}  \\
    \Xhline{2\arrayrulewidth}
    \end{tabular}
    }
    \caption{
    Performance on the blind, held-out test set of ShARC end-to-end task.
    }
    \label{tab:result-test}
\end{table*}

\section{Experiments}\label{sec.experi}

\subsection{Experimental Setup}\label{sec.exp_setup}

\paragraph{Dataset.}
ShARC \cite{saeidi-etal-2018-interpretation} dataset is the current benchmark to test entailment reasoning in conversational machine reading~\footnote{Leaderboard: \url{https://sharc-data.github.io/leaderboard.html}}.
The dataset contains 948 rule texts clawed from 10 government websites, in which 65\% of them are plain text with in-line conditions while the rest 35\% contain bullet-point conditions.
Each rule text is associated with a dialog tree (follow-up QAs) that considers all possible fulfillment combinations of conditions.
In the data annotation stage, parts of the dialogs are paraphrased into the user scenario.
% with some redundant information added to mimic the real-world situations. 
These parts of dialogs are marked as \texttt{evidence} which should be extracted (entailed) from the user scenario, and are not provided as inputs for evaluation.
The inputs to the system are the rule text, user question, user scenario, and dialog history (if any). The output is the answer among {Yes}, {No}, {Irrelevant}, or a follow-up question.
The train, development, and test dataset sizes are 21890, 2270, and 8276, respectively.

\paragraph{Evaluation Metrics.}
The decision making sub-task uses macro- and micro- accuracy of four classes ``Yes'', ``No'', ``Irrelevant'', ``Inquire'' as metrics. 
For the question generation sub-task, we evaluate models under both the official end-to-end setting \cite{saeidi-etal-2018-interpretation} and the recently proposed oracle setting \cite{gao-etal-2020-explicit}.
In the official setting, the BLEU score \cite{papineni-etal-2002-bleu} is calculated only when both the ground truth decision and the predicted decision are ``Inquire'', which makes the score dependent on the model's ``Inquire'' predictions.
For the oracle question generation setting, models are asked to generate a question when the ground truth decision is ``Inquire''.

% \paragraph{Baselines.}
% We compare with the following baseline models.
% \begin{itemize}
% \setlength{\itemsep}{0pt}
% \setlength{\parsep}{0pt}
% \setlength{\parskip}{0pt}
%     \item Seq2Seq \cite{saeidi-etal-2018-interpretation} formulates this problem as a generative question answering via a LSTM-based encoder-decoder model. 
%     \item Pipeline \cite{saeidi-etal-2018-interpretation} is a pipeline baseline which combines the random forest decision classification, TFIDF based logistic regression model for entailment, and rule-based model for question generation.
%     \item BERTQA \cite{zhong-zettlemoyer-2019-e3} is a BERT-based extractive question answering model which directly extracts the answer from the input sequence.
%     \item UrcaNet \cite{Sharma2019NeuralCQ} is a extension of BERTQA model with scenario, history and turn marker embeddings for decision making.
%     \item BiSon \cite{lawrence-etal-2019-attending} pre-trains a bi-directional sequence generation model with special placeholder tokens on ShARC.
%     \item E${}^{3}$ \cite{zhong-zettlemoyer-2019-e3} proposes an entailment-driven extraction approach to retrieve the unknown span, and edits it into a follow-up question.
%     \item EMT \cite{gao-etal-2020-explicit} explicitly tracks whether conditions listed in the rule text have already been satisfied for decision making, and adopts a coarse-to-fine reasoning strategy for question generation.
%     \item EMT+entailment \cite{gao-etal-2020-explicit} extends EMT by adding entailment prediction as multi-task learning.
% \end{itemize}

\paragraph{Implementation Details.}
% experimental setting
For the decision making sub-task, we finetune RoBERTa-base model \cite{Wolf2019HuggingFacesTS} with Adam \cite{adam} optimizer for 5 epochs with a learning rate of 5e-5, a warm-up rate of 0.1, a batch size of 16, and a dropout rate of 0.35. 
The number of inter-sentence transformer layers $L$ and the loss weight $\lambda$ for entailment prediction are hyperparameters.
We try 1,2,3 for $L$ and 1.0, 2.0, 3.0, 4.0, 5.0 for $\lambda$, and find the best combination is $L=2, \lambda=3.0$, based on the development set results.
For the question generation sub-task, we train a RoBERTa-base model to extract spans under the same training scheme above, and finetune UniLM \cite{unilm} 20 epochs for question rephrasing with a batch size of 16, a learning rate of 2e-5, and a beam size 10 for decoding in the inference stage.
We repeat 5 times with different random seeds for all experiments on the development set and report the average results along with their standard deviations.
It takes two hours for training on a 4-core server with an Nvidia GeForce GTX Titan X GPU.

\subsection{Results}
\paragraph{Decision Making Sub-task.}
The decision making results in macro- and micro- accuracy on the blind, held out test set of ShARC are shown in Table \ref{tab:result-test}. \modelnameshort outperforms the previous best model EMT \cite{gao-etal-2020-explicit} by 3.8\% in micro-averaged accuracy and 3.5\% in macro-averaged accuracy.
We further analyze the class-wise decision prediction accuracy on the development set of ShARC in Table \ref{tab:result-clf}, and find that \modelnameshort have far better predictions than all existing approaches whenever a decision on the user's fulfillment is needed (``Yes'', ``No'', ``Inquire''). It is because the predicted decisions from \modelnameshort are made upon the predicted entailment states while previous approaches do not build the connection between them.

\begin{table}[!t]
    \small
    \centering
    \resizebox{0.85\columnwidth}{!}{
    \begin{tabular}{l | c c c c }
    \Xhline{2\arrayrulewidth}
  Models & {Yes} & {No} & {Inq.} & {Irr.}  \\
    \hline
    \hline
    BERTQA    & 61.2 & 61.0  &  62.6 & 96.4       \\
    E${}^{3}$ & 65.9 & 70.6 & 60.5 & 96.4   \\
    UrcaNet & 63.3 & 68.4 & 58.9 & 95.7   \\
    EMT   & {70.5} & {73.2} & {70.8} & {98.6}  \\
    \modelnameshort   & \textbf{71.9} & \textbf{75.8} & \textbf{73.3} & \textbf{99.3}  \\
    \Xhline{2\arrayrulewidth}
    \end{tabular}
    }
    \caption{
    Class-wise decision prediction accuracy among ``Yes'', ``No'', ``Inquire'' and ``Irrelevant'' on the development set of ShARC.
    }
    \label{tab:result-clf}
\end{table}

\paragraph{Question Generation Sub-task.}
\modelnameshort outperforms existing methods under both the official end-to-end setting (Table \ref{tab:result-test}) and the recently proposed oracle setting (Table \ref{tab:result-dev-qg}).
Because the comparison among models is only fair under the oracle question generation setting \cite{gao-etal-2020-explicit}, we compare \modelnameshort with E${}^{3}$ \cite{zhong-zettlemoyer-2019-e3}, E${}^{3}$+UniLM \cite{gao-etal-2020-explicit}, EMT \cite{gao-etal-2020-explicit}, and our ablation \modelnameshort (BERT) in Table \ref{tab:result-dev-qg}.
Interestingly, we find that, in this oracle setting, our proposed simple approach is even better than previous sophisticated models such as E${}^{3}$ and EMT which jointly learn question generation and decision making via multi-task learning.
From our results and investigations, we believe the decision making sub-task and the follow-up question generation sub-task do not share too many commonalities so the results are not improved for each task in their multi-task training. 
On the other hand, our question generation model is easy to optimize because this model is separately trained from the decision making one, which means there is no need to balance the performance between these two sub-tasks.
Besides, RoBERTa backbone performs comparably with its BERT counterpart.

In our detailed analyses, we find \modelnameshort can locate the next questionable sentence with 77.2\% accuracy, which means \modelnameshort utilizes the user scenario and dialog history well to locate the next underspecified condition.
We try to add entailment prediction supervision to help \modelnameshort to locate the unfulfilled condition but it does not help.
We also try to simplify our approach by directly finetuning UniLM to learn the mapping between concatenated input sequences and the follow-up clarification questions. However, the poor result (around 40 for BLEU1) suggests this direction still remains further investigations.

\begin{table}[!t]
    \centering
    \small 
    \resizebox{0.95\columnwidth}{!}{
    \begin{tabular}{l | c c  }
    \Xhline{2\arrayrulewidth}
    Models &  BLEU1 & BLEU4  \\
    \hline\hline
    E${}^{3}$   & 52.79\scriptsize{$\pm$2.87} & 37.31\scriptsize{$\pm$2.35}  \\
    E${}^{3}$+UniLM  & 57.09\scriptsize{$\pm$1.70} & 41.05\scriptsize{$\pm$1.80}  \\
    EMT  & {62.32}\scriptsize{$\pm$1.62} & {47.89}\scriptsize{$\pm$1.58} \\
    \modelnameshort (BERT) & {64.13}\scriptsize{$\pm$0.43} & {50.73}\scriptsize{$\pm$0.72} \\
    \modelnameshort & \textbf{64.23}\scriptsize{$\pm$0.84} & \textbf{50.85}\scriptsize{$\pm$0.89} \\
    \Xhline{2\arrayrulewidth}
    \end{tabular}
    }
    \caption{
    Oracle question generation performance on the development set of ShARC. 
    }
    \label{tab:result-dev-qg}
\end{table}

\subsection{Ablation Study}

Table \ref{tab:result-ablation} shows an ablation study of \modelnameshort for the decision making sub-task on the development set of ShARC, and we have the following observations:

\paragraph{RoBERTa \textit{vs.} BERT.} \modelnameshort (BERT) replaces the RoBERTa backbone with BERT while other modules remain the same. 
The better performance of RoBERTa backbone matches findings from \citet{Talmor2019oLMpicsO}, which indicate that RoBERTa can capture negations and handle conjunctions of facts better than BERT.
% The performance is slightly worse since RoBERTa can capture negations and handle conjunctions of facts better than BERT \cite{Talmor2019oLMpicsO}.

\begin{table}[!t]
    \small
    \centering
    \resizebox{1.0\columnwidth}{!}{
    \begin{tabular}{l | c c }
    \Xhline{2\arrayrulewidth}
    Models & Micro Acc. & Macro Acc.  \\
    \hline
    \hline
    \modelnameshort & {74.97}\scriptsize{$\pm$0.27} & {79.55}\scriptsize{$\pm$0.35}  \\
    \modelnameshort (BERT)  & 73.07\scriptsize{$\pm$0.21} & 77.77\scriptsize{$\pm$0.24} \\
    \modelnameshort (w/o EDU)  & 73.34\scriptsize{$\pm$0.22} & 78.25\scriptsize{$\pm$0.57} \\
    \modelnameshort (w/o Trans)  & 74.25\scriptsize{$\pm$0.36} & 78.78\scriptsize{$\pm$0.57} \\
    \modelnameshort (w/o $\mathbf{\tilde{e}}$)  & 73.55\scriptsize{$\pm$0.26} & 78.19\scriptsize{$\pm$0.30} \\
    \modelnameshort (w/o $\mathbf{V}_{\text{EDU}}$)  & 72.95\scriptsize{$\pm$0.23} & 77.53\scriptsize{$\pm$0.19} \\
    \Xhline{2\arrayrulewidth}
    \end{tabular}
    }
    \caption{
    Ablation Study of \modelnameshort for decision making on the development set of ShARC. 
    }
    \label{tab:result-ablation}
\end{table}

\paragraph{Discourse Segmentation \textit{vs.} Sentence Splitting.} \modelnameshort (w/o EDU) replaces the discourse segmentation based rule parsing with simple sentence splitting, and we observe there is a 1.63\% drop on the micro-accuracy. This is intuitive because we observe 65\% of the rule texts in the training set contains in-line conditions. 
% If we do not segment out these in-line conditions, one rule sentence may contain several conditions like the rule text in Figure \ref{fig:model}, which would make the entailment state of that sentence ambiguous to define.
% Discern: macro 79.21, micro 75.75
% sentence: macro 76.62, micro 70.98
% evaluated on 1027/2270 examples
To better understand the effect of discourse segmentation, we also evaluate \modelnameshort and \modelnameshort (w/o EDU) on just that portion of examples that contains multiple EDUs. The micro-accuracy of decision making is 75.75 for \modelnameshort while it is 70.98 for \modelnameshort (w/o EDU). The significant gap shows that discourse segmentation is extremely helpful.
% for rule texts which have in-line conditions.

\paragraph{Are Inter-sentence Transformer Layers Necessary?} 
We investigate the necessity of inter-sentence transformer layers because RoBERTa-base already has 12 transformer layers, in which the sentence-level \texttt{[CLS]} representations can also interact with each other via multi-head self-attention.
Therefore, we remove the inter-sentence transformer layers and use the RoBERTa encoded \texttt{[CLS]} representations for entailment prediction and decision making.
The results show that removing the inter-sentence transformer layers (\modelnameshort w/o Trans) hurts the performance, which suggests that the inter-sentence self-attention is essential.

\paragraph{Both Condition Representations and Entailment Vectors Facilitate Decisions.} 
We remove either the condition representations $\mathbf{\tilde{e}}_i$ or the entailment vectors $\mathbf{V}_{\text{EDU}}$ in Eqn.\ref{eqn:alpha} \& \ref{eqn:summary} for decision predictions.
The results show that both sides of the information are useful for making decisions. Presumably, the condition representations account for the logical forms of rule texts and entailment vectors contain the fulfillment states for these conditions.

\subsection{Analysis of Logical Structure of Rules}

To see how \modelnameshort understands the logical structure of rules, we evaluate the decision making accuracy according to the logical types of rule texts.
Here we define four logical types: ``Simple'', ``Conjunction'', ``Disjunction'', ``Other'', which are inferred from the associated dialog trees.
``Simple'' means there is only one requirement in the rule text while ``Other'' denotes the rule text have complex logical structures, for example, a conjunction of disjunctions or a disjunction of conjunctions.
Table \ref{tab:result-logic} shows decision prediction results categorized by different logical structures of rules.
\modelnameshort achieves the best performance on the ``Simple'' logical type which only needs to determine the single condition is satisfied or not.
On the other hand, \modelnameshort does not perform well on rules in the format of disjunctions. We conduct further analysis on this category and find that the error comes from user scenario interpretation: the user has already provided his fulfillment in the user scenario but \modelnameshort fails to extract it.
Detailed analyses are further conducted in the following section.

\begin{table}[!t]
    % \small
    \centering
    \resizebox{1.0\columnwidth}{!}{
    \begin{tabular}{l | c c c }
    \Xhline{2\arrayrulewidth}
  Logical Type & \# samples & Micro Acc. & Macro Acc.  \\
    \hline
    \hline
    Simple    & 569 &  82.78\scriptsize{$\pm$1.48} & 86.91\scriptsize{$\pm$1.31}        \\
    Disjunction & 726 & 69.97\scriptsize{$\pm$1.85} & 75.89\scriptsize{$\pm$1.38}   \\
    Conjunction & 698 & 74.47\scriptsize{$\pm$2.41} & 79.78\scriptsize{$\pm$1.74}   \\
    Other & 277 & 73.29\scriptsize{$\pm$2.53} & 77.17\scriptsize{$\pm$1.54}   \\
    \Xhline{2\arrayrulewidth}
    \end{tabular}
    }
    \caption{
    Decision prediction accuracy categorized by logical types of rules on the ShARC development set.
    }
    \label{tab:result-logic}
\end{table}

\begin{table*}[!t]
    \small
    \centering
    \resizebox{0.9\textwidth}{!}{
    \begin{tabular}{l | c c c c }
    \Xhline{2\arrayrulewidth}
  \multirow{2}{*}{Dataset} &  \multicolumn{2}{c}{Decision Making} &  \multicolumn{2}{c}{Entailment Prediction} \\  
     & Micro Acc. & Macro Acc. & Micro Acc. & Macro Acc.  \\
    \hline
    \hline
    ShARC (Answerable)    & 73.55\scriptsize{$\pm$0.33} & 73.46\scriptsize{$\pm$0.27}  & 86.41\scriptsize{$\pm$0.39} & 81.13\scriptsize{$\pm$0.39} \\
    ~~~~ Dialog History Subset & 79.29\scriptsize{$\pm$1.62} & 76.37\scriptsize{$\pm$1.95} & 92.41\scriptsize{$\pm$0.38} & 90.12\scriptsize{$\pm$0.68}   \\
    ~~~~ Scenario Subset & 63.50\scriptsize{$\pm$1.58} & 60.18\scriptsize{$\pm$1.72} & 82.76\scriptsize{$\pm$0.46} & 59.40\scriptsize{$\pm$1.04}  \\
    ShARC (Evidence) & 84.93\scriptsize{$\pm$0.29} & 84.37\scriptsize{$\pm$0.24} & 91.46\scriptsize{$\pm$0.68} & 89.90\scriptsize{$\pm$1.40}  \\
    \Xhline{2\arrayrulewidth}
    \end{tabular}
    }
    \caption{
    Decision making and entailment prediction results over different subsets of the ShARC development set.
    }
    \label{tab:result-imp}
\end{table*}

\subsection{How Far Has the Problem Been Solved?} \label{sec:SubsetPerformance}
In order to figure out the limitations of \modelnameshortnsp, and the current challenges of ShARC CMR, 
we disentangle the challenges of scenario interpretation and dialog understanding in ShARC by selecting different subsets, and evaluate decision making and entailment prediction accuracy on them.

\paragraph{Baseline.}
Because the classification for unanswerable questions (``irrelevant'' class) is nearly solved (99.3\% in Table \ref{tab:result-clf}), we create the baseline subset by removing all unanswerable examples from the development set. Results for this baseline are shown in \textbf{ShARC (Answerable)} of Table~\ref{tab:result-imp}.

\paragraph{Dialog History Subset.}
We first want to see how \modelnameshort understands dialog histories (follow-up QAs) without the influence of user scenarios. 
Hence, we create a subset of ShARC (Answerable) in which all samples have an empty user scenario. 
The performance over 224 such samples is shown in ``Dialog History Subset'' of Table \ref{tab:result-imp}. Surprisingly, the results on this portion of samples are much better than the overall results, especially for the entailment prediction (92.41\% micro-accuracy). 
% Here we believe our model can learn the connection between dialog question-answer pairs and the rule conditions well, leading to accurate entailment and decision predictions.

\paragraph{Scenario Subset.}
With the curiosity to see what is the bottleneck of our model, we test the model ability on scenario interpretation. 
Similarly, we create a ``Scenario Subset'' from ShARC (Answerable) in which all samples have an empty dialog history.
Results in Table \ref{tab:result-imp} (``Scenario Subset'') show that interpreting scenarios to extract the entailment information within is exactly the current bottleneck of \modelnameshortnsp.
We analyze 100 error cases on this subset and find that various types of reasoning are required for scenario interpretation, including numerical reasoning (15\%), temporal reasoning (12\%), and implication over common sense and external knowledge (46\%). 
Besides, \modelnameshort still fails to extract user's fulfillment when the scenarios paraphrase the rule texts (27\%).
Examples for each type of error are shown in Figure \ref{tab:error}.
Among three classes of entailment states, we find that \modelnameshort fails to predict \textsc{Entailment} or \textsc{Contradiction} precisely -- it predicts \textsc{Neutral} in most cases for scenario interpretation, resulting in high micro-accuracy in entailment prediction but the macro-accuracy is poor. The decision accuracy is subsequently hurt by the entailment results.

\paragraph{ShARC (Evidence).}
Based on the above observation, we replace the user scenario in the ShARC (Answerable) by its \texttt{evidence} and re-evaluate the overall performance on these answerable questions. As described in Section \ref{sec.exp_setup} Dataset, the \texttt{evidence} is the part of dialogs that should be entailed from the user scenario.
Table \ref{tab:result-imp} shows that the model improves 11.38\% in decision making micro-accuracy if no scenario interpretation is required, which validates our above observation.

\begin{figure*}[!t]
    \centering
    \footnotesize
    \resizebox{1.0\textwidth}{!}{
    \setlength\tabcolsep{0.8pt}
    \begin{tabular}{p{0.25\textwidth}p{0.05\textwidth}p{0.70\textwidth}}
    \toprule
        Error Type & \% & Example \\
    \midrule
        \multirow{5}{*}{Numerical Reasoning} & \multirow{5}{*}{15} & \textbf{Relevant Rule}: Each attachment must be less than 10MB.  \\
        & & \textbf{Scenario}: The attachment right now isn't less than 10MB, but I think I can compress so it becomes less than 10MB. \\
        & & \textbf{Question}: Can I upload the attachment? \\ 
        & & \textbf{Entailment State} ~~~ {Gold}: \texttt{Entailment}; ~~~ {Predict}: \texttt{Contradiction} \\ 
    \specialrule{0em}{1pt}{1pt}
    \hdashline
    \specialrule{0em}{1pt}{1pt}
        \multirow{6}{*}{Temporal Reasoning} & \multirow{6}{*}{12} & \textbf{Relevant Rule}: The Additional State Pension is an extra amount of money you could get on top of your basic State Pension if you’re: … * a woman born before 6 April 1953 \\
        & & \textbf{Scenario}: I live with my husband.  We both have worked all our lives.  Both of us were born in 1950. \\
        & & \textbf{Question}: Can I get Additional State Pension? \\ 
        & & \textbf{Entailment State} ~~~ {Gold}: \texttt{Entailment}; ~~~ Predict: \texttt{Contradiction} \\ 
    \specialrule{0em}{1pt}{1pt}
    \hdashline
    \specialrule{0em}{1pt}{1pt}
        \multirow{5}{*}{Commonsense Reasoning} & \multirow{5}{*}{46} & \textbf{Relevant Rule}: Homeowners may apply for up to \$200,000 to repair or replace their primary residence to its pre-disaster condition. \\
        & & \textbf{Scenario}: My home was flooded \\
        & & \textbf{Question}: Is this loan suitable for me? \\ 
        & & \textbf{Entailment State} ~~~ {Gold}: \texttt{Entailment}; ~~~ Predict: \texttt{Unknown} \\ 
    \specialrule{0em}{1pt}{1pt}
    \hdashline
    \specialrule{0em}{1pt}{1pt}
        \multirow{6}{*}{Paraphrase Reasoning} & \multirow{6}{*}{27} & \textbf{Relevant Rule}: The Montgomery GI Bill (MGIB) is an educational assistance program enacted by Congress to attract high quality men and women into the Armed Forces. \\
        & & \textbf{Scenario}: I applied and found out I can get a loan. My dad wants me to join the army, but I don't. I'd rather go to school. \\
        & & \textbf{Question}: Does this program meet my needs? \\ 
        & & \textbf{Entailment State} ~~~ {Gold}: \texttt{Contradiction}; ~~~ Predict: \texttt{Unknown} \\ 
    \bottomrule
    \end{tabular}
    }
    \caption{
        Types of scenario interpretation errors in the development data based on 100 samples.
    }
    % \vspace{-0.1in}
    \label{tab:error}
\end{figure*}

\section{Related Work}
\paragraph{Entailment Reasoning in Reading Comprehension.} 
Understanding entailments (or implications) of text is essential in dialog and question answering systems.
ROPES \cite{lin-etal-2019-reasoning} requires reading descriptions of causes and effects and applying them to situated questions, while ShARC \cite{saeidi-etal-2018-interpretation}, the focus of \modelnameshortnsp, requires to understand rules and apply them to questions asked by users in a conversational manner.
Most existing methods simply use BERT to classify the answer without considering the structures of rule texts \cite{zhong-zettlemoyer-2019-e3,Sharma2019NeuralCQ,lawrence-etal-2019-attending}. 
\citet{gao-etal-2020-explicit} propose Explicit Memory Tracker (EMT), which firstly addresses entailment-oriented reasoning. At each dialog turn, EMT recurrently tracks whether conditions listed in the rule text have already been satisfied to make a decision. 

In this paper, we also explicitly model entailment reasoning for decision making, but there are three key differences between our \modelnameshort and EMT:
(1) we apply discourse segmentation to parse the rule text, which is extremely helpful because there are many in-line conditions in rules;
(2) Our stacked inter-sentence transformer layers extract better features for entailment prediction, which could be seen as a generalization of their recurrent explicit memory tracker.
(3) Different from their utilization of entailment prediction which is treated as multi-task learning for decision making, we directly build the dependency between entailment prediction states and the predicted decisions.

\paragraph{Discourse Applications.}

Discourse analysis uncovers text-level linguistic structures (e.g., topic, coherence, co-reference), which can be useful for many downstream applications, such as coherent text generation \cite{bosselut-etal-2018-discourse} and text summarization \cite{joty-etal-2019-discourse,cohan-etal-2018-discourse,Xu2019DiscourseAwareNE}.
Recently, discourse information has also been introduced in neural reading comprehension. \citet{mihaylov-frank-2019-discourse} design a discourse-aware semantic self-attention mechanism to supervise different heads of the transformer by discourse relations and coreferring mentions.
Different from their use of discourse information, we use it as a parser  to segment surface-level in-line conditions for entailment reasoning.

\section{Conclusion}
In this paper, we present \modelnameshortnsp, a system that does discourse-aware entailment reasoning for conversational machine reading.
\modelnameshort explicitly builds the connection between entailment states of conditions and the final decisions.
Results on the ShARC benchmark shows that \modelnameshort outperforms existing methods by a large margin. 
We also conduct comprehensive analyses to unveil the limitations of \modelnameshort and challenges for ShARC.
In future, we plan to explore how to incorporate discourse parsing into the current decision making model for end-to-end learning. One possibility would be to frame them as multi-task learning with a common (shared) encoder. 
Another direction is leveraging current methods in question generation \cite{gao-etal-2019-interconnected,li-etal-2019-improving-question} to improve the follow-up question generation sub-task since \modelnameshort\ is on par with the previous best model EMT.

\section*{Acknowledgments}
We thank Max Bartolo and Patrick Lewis for evaluating our submitted models on the hidden test set.
The work described in this paper was partially supported by following projects from the Research Grants Council of the Hong Kong Special Administrative Region, China: CUHK 2300174 (Collaborative Research Fund, No. C5026-18GF); CUHK 14210717 (RGC General Research Fund).

\bibliographystyle{acl_natbib}
\bibliography{emnlp2020,anthology}

% \appendix

\end{document}